\documentclass[letterpaper, 10 pt, conference]{ieeeconf}

\IEEEoverridecommandlockouts                              

\let\labelindent\relax
\usepackage{enumitem}

\usepackage{array}
\usepackage{graphicx}
\usepackage{cite}
\usepackage{amsmath}
\usepackage{amssymb}
\usepackage{stmaryrd}
\usepackage{multirow}
\usepackage{algorithm}
\usepackage{algorithmic}
\usepackage[T1]{fontenc}
\usepackage{multicol}
\usepackage{stfloats}
\usepackage[caption = false]{subfig}
\usepackage{xcolor}

\usepackage{mathtools}

\usepackage{makecell}
\usepackage{scalerel}
\usepackage{array}
\usepackage{multirow}
\usepackage{dsfont}
\usepackage{url}

\newcolumntype{M}{>{\centering\arraybackslash}m{.2\textwidth}}
\newcolumntype{C}[1]{>{\centering\let\newline\\\arraybackslash\hspace{0pt}}p{#1}}
\newcolumntype{R}[1]{>{\raggedleft\let\newline\\\arraybackslash\hspace{0pt}}p{#1}}
\newcolumntype{L}[1]{>{\raggedright\let\newline\\\arraybackslash\hspace{0pt}}p{#1}}
\newcommand*\rot{\rotatebox{90}}
\newcommand\Tstrut{\rule{-3pt}{2.6ex}}       
\newcommand\Bstrut{\rule[-0.9ex]{-3pt}{0pt}} 
\newcommand{\TBstrut}{\rule{-3pt}{2.6ex} \rule[-0.9ex]{-2pt}{0pt}}  

\newcommand\mydots{\hbox to 1em{.\hss.\hss.}}

\definecolor{myblue}{RGB}{0, 250, 0}
\definecolor{mypink}{RGB}{237, 2, 140}

\usepackage[hidelinks]{hyperref}

\begin{document}

\title{\LARGE \bf
Learning Spatiotemporal Occupancy Grid Maps for Lifelong Navigation in Dynamic Scenes
}

\author{
Hugues Thomas$^{1}$
\quad
Matthieu Gallet de Saint Aurin$^{2}$
\quad 
Jian Zhang$^{3}$
\quad
Timothy D. Barfoot$^{1}$
\thanks{$^{1}$University of Toronto Institute for Aerospace Studies (UTIAS), Canada. $^{2}$ETH Z\"urich, Switzerland. $^{3}$Apple Inc. {\tt\small \{hugues.thomas, tim.barfoot\}@utoronto.ca, maaurin@student.ethz.ch, jianz@apple.com}.}
}

\maketitle


\IEEEpeerreviewmaketitle

\begin{abstract}

We present a novel method for generating, predicting, and using Spatiotemporal Occupancy Grid Maps (SOGM), which embed future information of dynamic scenes. Our automated generation process creates groundtruth SOGMs from previous navigation data. We build on prior work to annotate lidar points based on their dynamic properties, which are then projected on time-stamped 2D grids: SOGMs. We design a 3D-2D feedforward architecture, trained to predict the future time steps of SOGMs, given 3D lidar frames as input. Our pipeline is entirely self-supervised, thus enabling lifelong learning for robots. The network is composed of a 3D back-end that extracts rich features and enables the semantic segmentation of the lidar frames, and a 2D front-end that predicts the future information embedded in the SOGMs within planning. We also design a navigation pipeline that uses these predicted SOGMs. We provide both quantitative and qualitative insights into the predictions and validate our choices of network design with a comparison to the state of the art and ablation studies.

\end{abstract}



\section{Introduction}
\label{sec:1}

Robot navigation, and particularly local planning, is complex in the presence of dynamic obstacles. Anticipating future motions has begun to be possible with recent advances in machine learning and robotics. In our work, we aim at providing an accurate prediction of the temporal evolution of the environment, enabling a robot to proactively plan efficient and safe trajectories. As a lifelong learning approach, Reinforcement Learning (RL) often replaces the whole navigation pipeline; instead, we advocate the use of Self-Supervised Learning (SSL) to solve specific tasks helping robot navigation. We propose a method to predict the future motions of dynamic obstacles, which can easily be used in the navigation pipeline, along with proven localization and planning algorithms.

\begin{figure}[t]
    \centering
    \includegraphics[width=0.95\columnwidth, keepaspectratio=true]{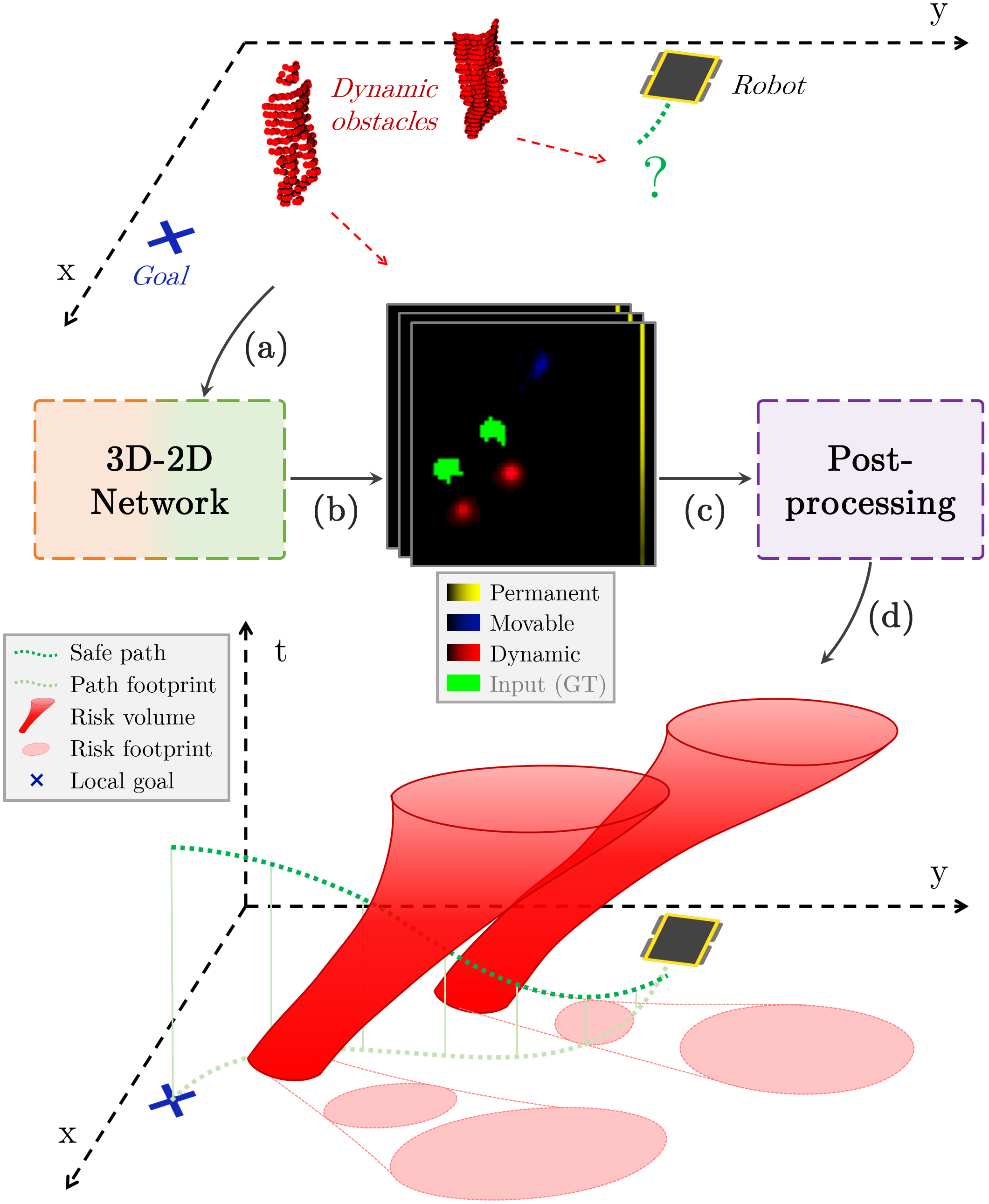}
    \vspace{-2ex}
    \caption{The robot faces dynamic obstacles (a). Our 3D-2D network forecasts their future as Spatiotemporal Occupancy Grid Maps (b), which are processed into Spatiotemporal Risk Maps (c). A planned trajectory is optimized in the 3D space $(xyt)$ (d).}
    \label{fig_intro}
    \vspace{-4ex}
\end{figure}

Building on our previous work \cite{thomas2021self}, we use a similar automated annotation process with mapping and ray-tracing algorithms, providing four semantic labels to individual lidar points: \textit{ground}, \textit{permanent} (structures such as walls), \textit{movable} (still-but-movable obstacles such as chairs), and \textit{dynamic} (such as people). In addition, we propose a Spatiotemporal Occupancy Grid Maps (SOGM) generation algorithm, converting the annotated lidar point clouds to SOGMs with three channels, one for each obstacle label. We then train a novel 3D-2D feedforward architecture to predict the future time steps of SOGMs, given past 3D lidar frames as input. As shown in Figure \ref{fig_intro}, the predicted SOGMs can be processed and used by an existing trajectory planning algorithm.

We define a SOGM as a 3D grid map, where each grid cell, independent from the others, contains an occupancy probability for a given position and time, in world coordinates. It can be interpreted as a set of 2D Occupancy Grid Maps (OGM), one for each future time step. Note that this is different from the Dynamic Occupancy Grid Maps used by \cite{schreiber2019long}, which are OGMs with additional velocity features in each cell. 
Therefore, the first crucial characteristic of our method is to be \textbf{point-centric}. As opposed to object-centric methods, we remain closer to the raw sensor data and inherently allow multi-modal predictions. A second characteristic is to include 3 types of obstacles as channels of the SOGM. Even if we are only interested in the prediction for the \textit{dynamic} obstacles, it allows a significant feature: our network can take into account interactions when predicting future motions. For example, dynamic obstacles will bounce off walls, avoid objects, or pass through open doors.

Our novel 3D-2D feedforward network architecture, shown in Figure \ref{fig_net}, is able to predict SOGMs from 3D lidar frames. The network input consists of consecutive frames, aligned on a global map and merged, to provide both spatial and temporal information. Our network starts with a 3D back-end, processing the lidar frames with KPConv layers \cite{thomas2019kpconv} to extract rich features that can be projected on a 2D grid or used for point cloud semantic segmentation. The 2D grid features are processed by a 2D front-end network outputting a SOGM. The characteristic of this 2D front-end is to be feedforward, without any recurrent connections. The occupancy predictions at consecutive time steps are computed by successive convolutions with independent weights. This enables our network to learn and predict a wider range of spatiotemporal patterns at different future horizons. The main contributions of this paper are:
\begin{itemize}[noitemsep, topsep=2pt, parsep=2pt, partopsep=2pt]
    \item The automated SOGM generation method.
    \item The novel 3D-2D feedforward neural network predicting three-channel SOGMs.
\end{itemize}

Additionally, we integrated SOGMs in an overall navigation stack. First, they are converted into Spatiotemporal Risk Maps (SRM), to reflect the risk of collision, linearly decreasing with the distance to occupied space. The SRMs are then used by a modified Timed Elastic Band (TEB) \cite{rosmann2015planning, rosmann2017integrated} local planner. TEB normally optimizes a trajectory in a three-dimensional spatiotemporal space $\left(x,y,t\right)$, maximizing the distance to discrete obstacles, but was adapted to minimize the risk value from the SRMs instead. We also keep the triaging idea from \cite{thomas2021self}, with some improvements. We reduce the delay in localization by using the raw lidar frames to locate against the map and keep the advantage of triaging by updating the map only with classified frames. While this paper focuses on the generation and prediction of SOGMs, we show early results of how we plan to use them in this overall navigation pipeline.



\section{Related Work}
\label{sec:2}

Navigation around dynamic obstacles has been studied for a long time in robotics. In terms of predicting obstacles' future motions, \cite{ziebart2009planning, kitani2012activity} learned a distribution of possible pedestrian trajectories using Inverse Optimal Control to recover a set of agent preferences consistent with prior demonstration. Following these preliminary works, several directions have been followed for dynamic obstacle forecasting.

\textbf{Object tracking and trajectory prediction} is a very common approach to enable motion planning in environments with dynamic obstacles. Following the success of recurrent neural networks (RNNs) and in particular long short-term memory networks (LSTMs) for trajectory prediction \cite{alahi2016social, gupta2018social}, the idea of isolating each obstacle as a distinct object has received a lot of attention. \cite{katyal2020intent} uses an LSTM-based network to predict people trajectories and plan around them, while \cite{peddi2020data} exploits a Hidden Markov Model to predict future states from a history of observations. Similarly, \cite{sathyamoorthy2020frozone} detects individual obstacles and predicts their speeds to avoid ``freezing zones''. Such object-centric methods are also used in the context of autonomous driving \cite{luo2018fast, casas2018intentnet}. However, they all rely on the detection and tracking results and do not easily incorporate multi-modal predictions, problems we do not face with our point-centric approach. Closer to our work, \cite{jain2020discrete} predicts future human motions as 2D heat maps, implicitly handling multi-modality like us, but is still relying on object-level predictions, and is also limited to 2D inputs, where our method leverages 3D features.

\textbf{Reinforcement Learning} has also been used extensively in recent years to replace standard motion planning algorithms \cite{long2018towards, liang2020crowdsteer, sathyamoorthy2020densecavoid, everett2021collision, strudel2020learning, liu2020robot}. However, standard local planners have proven to be very reliable methods, especially when it comes to producing dynamically feasible trajectories, which most RL methods fail to do. Even when the feasibility is ensured \cite{patel2021dwa}, the whole planning algorithm is embedded into a black box end-to-end neural network, which is difficult to tune, debug and interpret. We chose to keep a standard local planner, with its guarantees and interpretability, and use a self-supervised deep learning method to predict the future motion of the dynamic obstacles.

\textbf{OGM prediction} approaches are the closest to our work, and can be separated in two groups either using handcrafted or learned features. Handcrafted approaches usually rely on a model for human motion prediction. \cite{pierson2019dynamic} predicts a Dynamic Risk Density based on the occupancy density and velocity field of the environment. \cite{huang2020safe} extends this idea with a Gaussian Process regulated risk map of tracked pedestrians and cars.  Other recent works focus on adapting the uncertainty of the predictive model \cite{fisac2018probabilistically, bajcsy2019scalable, bansal2020hamilton}, using real-time Bayesian frameworks in a closed-loop on real data. These methods either rely on object detection and tracking or are based on instantaneous velocities, and not able to predict future locations of obstacles accurately. 
Learned methods, usually based on video frame prediction architectures \cite{lotter2016deep, wang2018predrnn++, wang2019memory}, are better at predicting complex futures. \cite{mohajerin2019multi} introduces a difference-learning-based recurrent architecture to extract and incorporate motion information for OGM prediction. \cite{schreiber2020motion} presents an LSTM-based encoder-decoder framework to predict the future environment represented as Dynamic Occupancy Grid Maps. They introduce recurrent skip connections into the network to reduce prediction blurriness. To account for the spatiotemporal evolution of the occupancy state of both static object and dynamic objects, \cite{toyungyernsub2020double} proposes a double-prong network architecture.

However, two major differences remain with our approach. First, these methods all take previous OGMs as input, effectively losing valuable information in the shape patterns that common 3D sensors can capture. To our knowledge, we are the first to fill this gap in the literature, by incorporating 3D features in the forecasting of OGMs with the 3D backbone of our network. Second, we are also the first, to our knowledge, to predict a sequence of OGMs without recurrent layers. We argue feedforward architectures are easier to train and understand. Eventually, our network is able to make a distinction between different semantic classes, leveraging interactions between them, when predicting future occupancy.


\section{Risk-Aware Navigation System}
\label{sec:3}

This section describes our approach, from the automated SOGM generation process, to the SOGM prediction with our novel 3D-2D feedforward architecture, and including training and inference details. Eventually, we present our risk-aware navigation system, with a modified TEB planner using SRMs. We encourage reproducibility and application to new datasets with our open-source implementation\footnote{\urlstyle{sf}\textcolor{mypink}{\url{https://github.com/utiasASRL/Deep-Collison-Checker}}}.

\subsection{Automated SOGM Generation}

Generating the groundtruth SOGMs automatically allows us to train the network without human intervention, in a self-supervised manner. We follow the same lifelong learning principles as \cite{thomas2021self}, allowing our network to learn from new situations, encountered through the robot's life.

\begin{figure}[t]
    \centering
    \includegraphics[width=0.999\columnwidth, keepaspectratio=true]{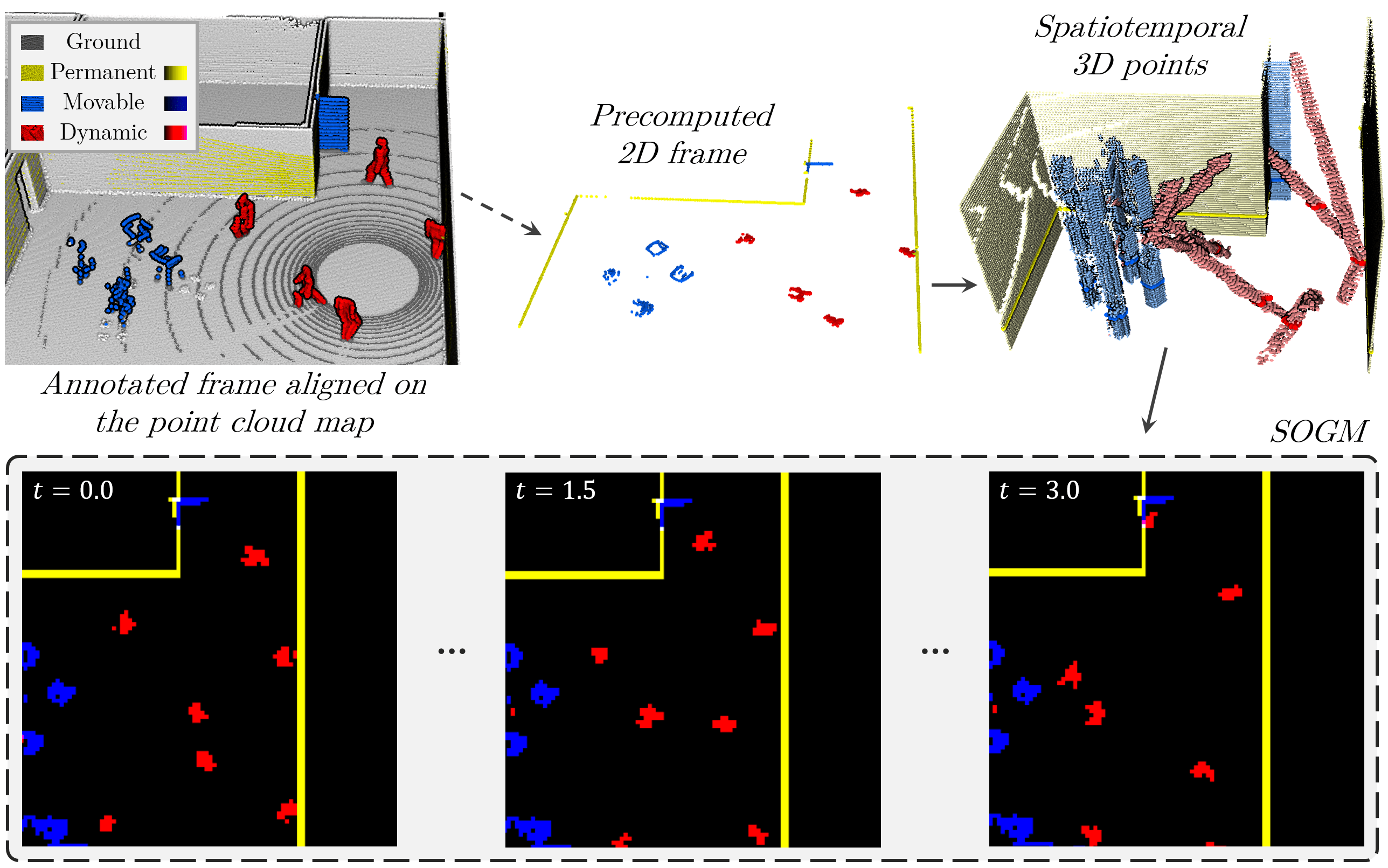}
    \caption{During pre-processing, every frame is semantically filtered and projected in 2D. During training, the 2D frames are stacked in 3D according to their timestamps and projected to a 3D grid to create the SOGMs.}
    \label{fig_gen}
\end{figure}

We first annotate lidar frames, using the method from \cite{thomas2021self}. A combination of a point cloud SLAM algorithm and a point cloud ray-tracing algorithm is used to create and then annotate a point cloud map for each previous navigation session.  The four semantic labels, \textit{ground}, \textit{permanent}, \textit{movable}, and \textit{dynamic}, are identified on the map and projected back to the lidar frames of each session, which will be used to train the 3D semantic segmentation part of our network.

We can then proceed to the generation of SOGMs with the three obstacles classes: \textit{permanent}, \textit{movable}, and \textit{dynamic}. To avoid the problem of rotating grids, we precompute intermediate 2D point clouds, which can easily be rotated and then transformed into SOGMs during training (Figure \ref{fig_gen}). In addition such a sparse structure in lighter to save than full grids. First, each annotated lidar frame is filtered to eliminate the ground points. We remove every point that is not annotated as an obstacle class, or that is closer than $20$ cm from the ground plane. Then we project all the points on the ground plane and subsample the obtained 2D point cloud with a grid size of $3$cm.

At training time, the 2D point clouds we need are loaded and stacked along a third dimension according to their timestamp. After being rotated for data augmentation, this spatiotemporal point cloud is projected to a SOGM structure of spatial resolution $dl_\mathrm{2D}=12$cm and temporal resolution $dt=0.1$s. The \textit{permanent} and \textit{movable} occupancy from all time steps of the SOGM are merged because they are not moving. Therefore, in addition to the future locations of dynamic obstacles, our network also learns to complete partially seen static objects.

\begin{figure}[b]
    \vspace{-2ex}
    \centering
    \includegraphics[width=0.999\columnwidth, keepaspectratio=true]{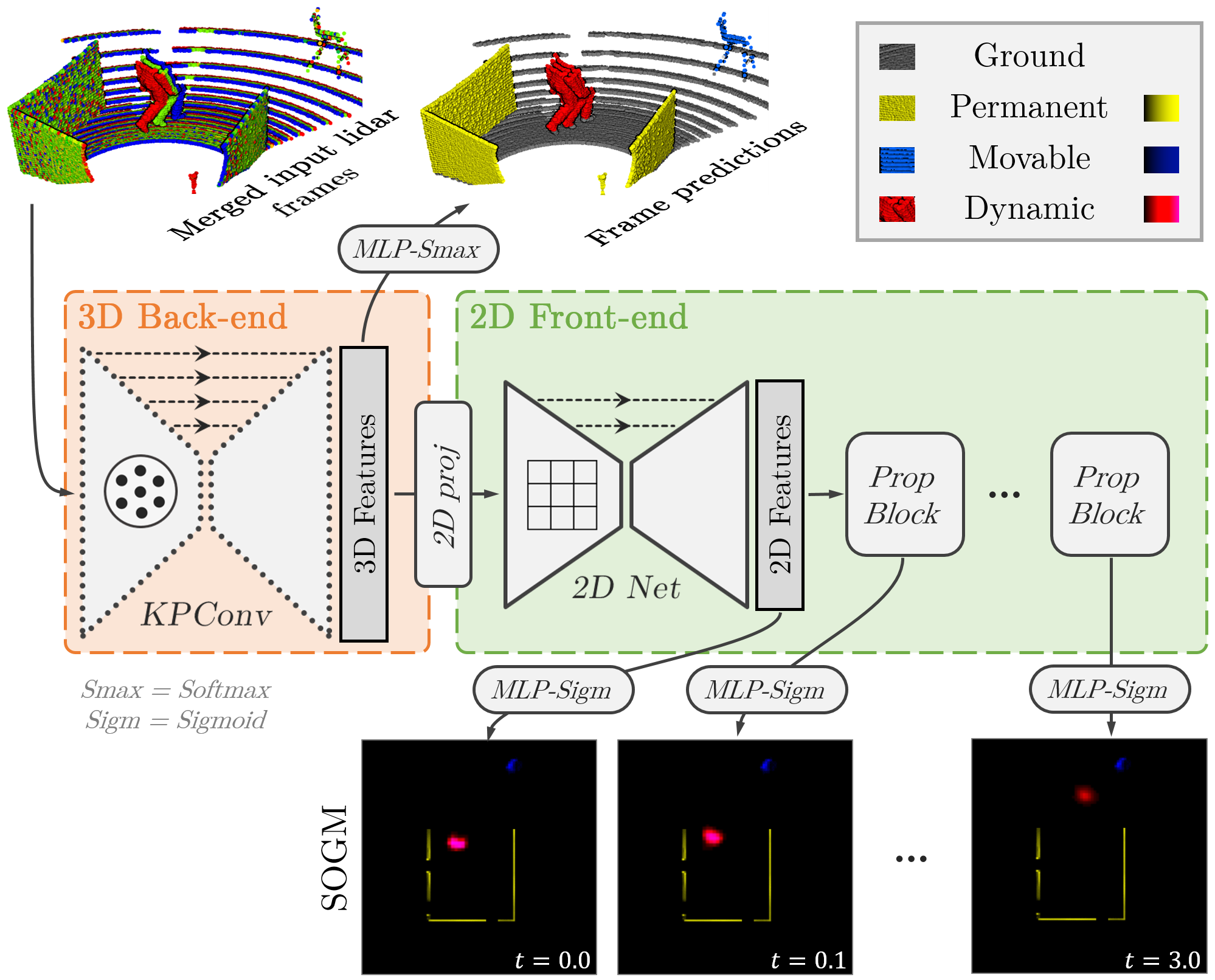}
    \vspace{-2ex}
    \caption{Illustration of our 3D-2D feedforward architecture. The 3D back-end is a 5-layer KPConv architecture, producing features at the point level. The 2D front-end is composed of a 3-layer 2D U-Net architecture, followed by consecutive convolutional propagation blocks.}
    \label{fig_net}
\end{figure}

\subsection{Network Architecture for SOGM Prediction}

Our network architecture (Figure \ref{fig_net}) is composed of two parts, a 3D back-end, and a 2D front-end. The 3D back-end is a KPConv network \cite{thomas2019kpconv} predicting a semantic label for each input point, that we use in our improved navigation system (see section \ref{sec:3.4}). Predicting 3D labels helps the network training by providing an additional supervisory signal and ensures that rich features are passed to the 2D front-end. We keep the KP-FCNN architecture and parameters of the original paper: a U-Net with five levels, each composed of two ResNet layers. The network input is a point cloud made from $n_\mathrm{f}=3$ lidar frames aligned in the map coordinates and merged. We only keep the points inside a $R_\mathrm{in}=8$m radius, as we are interested in the local vicinity of the robot. Each point is assigned a one-hot $n_\mathrm{f}$-dimensional feature vector, encoding of the lidar frame to which it belongs. Following the KPConv paper, we control the input point cloud density with a grid subsampling ($dl_\mathrm{3D}=6$cm).

The 3D point features are passed to the 2D front-end with a grid projection using the same spatial resolution $dl_\mathrm{2D}$ as the SOGM. The size of the grid is determined as the inscribed square in the $R_\mathrm{in}$-radius circle: $h_\mathrm{grid} = w_\mathrm{grid} = 94$. Features from points located in the same cell are averaged to get the new cell features. The features of the empty cells are set to zero. The obtained 2D feature map is then processed by an image U-Net architecture with three levels, each composed of two ResNet layers, to diffuse the information contained in sparse locations to the whole grid. This dense feature map is used to predict the initial time step of the SOGM. Then, it is processed by successive propagation blocks, each composed of two ResNet layers. The output of each propagation block is used to predict the corresponding time step of the SOGM. We define the final prediction time $T=3.0$s, meaning that our SOGMs have  $n_T = T / dt + 1 = 31$ time steps in total. Note that the \textit{permanent} and \textit{movable} predictions are redundant but we keep them to force the network to keep the knowledge of their location after the propagation blocks. It helps to learn interactions between the classes further into the future.

\subsection{Network Training and Inference}

We define a particular training loss for our network:

\begin{equation}
\label{eq2}
    L_\mathrm{tot} = \lambda_1 L^\mathrm{3D} + \lambda_2 \sum\limits_{k<n_T}{  \frac{L^\mathrm{2D}_k}{n_T}       }
\end{equation}

\noindent where $\lambda_1=1.0$, $\lambda_2=10.0$, $L^\mathrm{3D}$ is the standard cross entropy loss for semantic segmentation with a KPConv network, and $L^\mathsf{2D}_k$ is the loss applied to layer $k$ of our SOGM predictions:

\begin{equation}
\label{eq3}
    L^\mathrm{2D}_k = \sum\limits_{i \in M_k}{{BCE}(x_{k,i}, y_{k,i})}
\end{equation}


\noindent where $x_{k,i}$ is the network logit at the pixel $i$ of the time-step layer $k$ in the SOGM, $y_{k,i}$ is its corresponding groundtruth and $BCE$ is a Binary Cross-Entropy loss. Note that for clarity, we use a simple index $i$ for 2D pixels. The SOGM loss is thus a masked Binary Cross-Entropy, where the mask $M_k$ is here to help ignore the over-represented empty pixels and focus on the positive examples. We first tried a mask covering the positive groundtruth values in addition to some random pixels (GT-Mask), but then improved it to cover the union of positive groundtruth and positive prediction pixels (Active-Mask) to help reduce the false positives (See Figure \ref{fig_mask} bottom).

During training, we only use rotation augmentation around the vertical axis. The rest of the training parameters are kept identical as in the original KPConv paper \cite{thomas2019kpconv}. In this setup, our input point clouds contain on average $20,000$ points. We use a variable batch size targeting $B=6$, for an average of $85,000$ points per batch. More details can be found in our open-source code.

\subsection{Risk-Aware Navigation}
\label{sec:3.4}

Although our navigation system is not fully evaluated in this paper, we find it important to describe it here, to place some context around our other major contributions.

\begin{figure}[t]
    \centering
    \includegraphics[width=0.999\columnwidth, keepaspectratio=true]{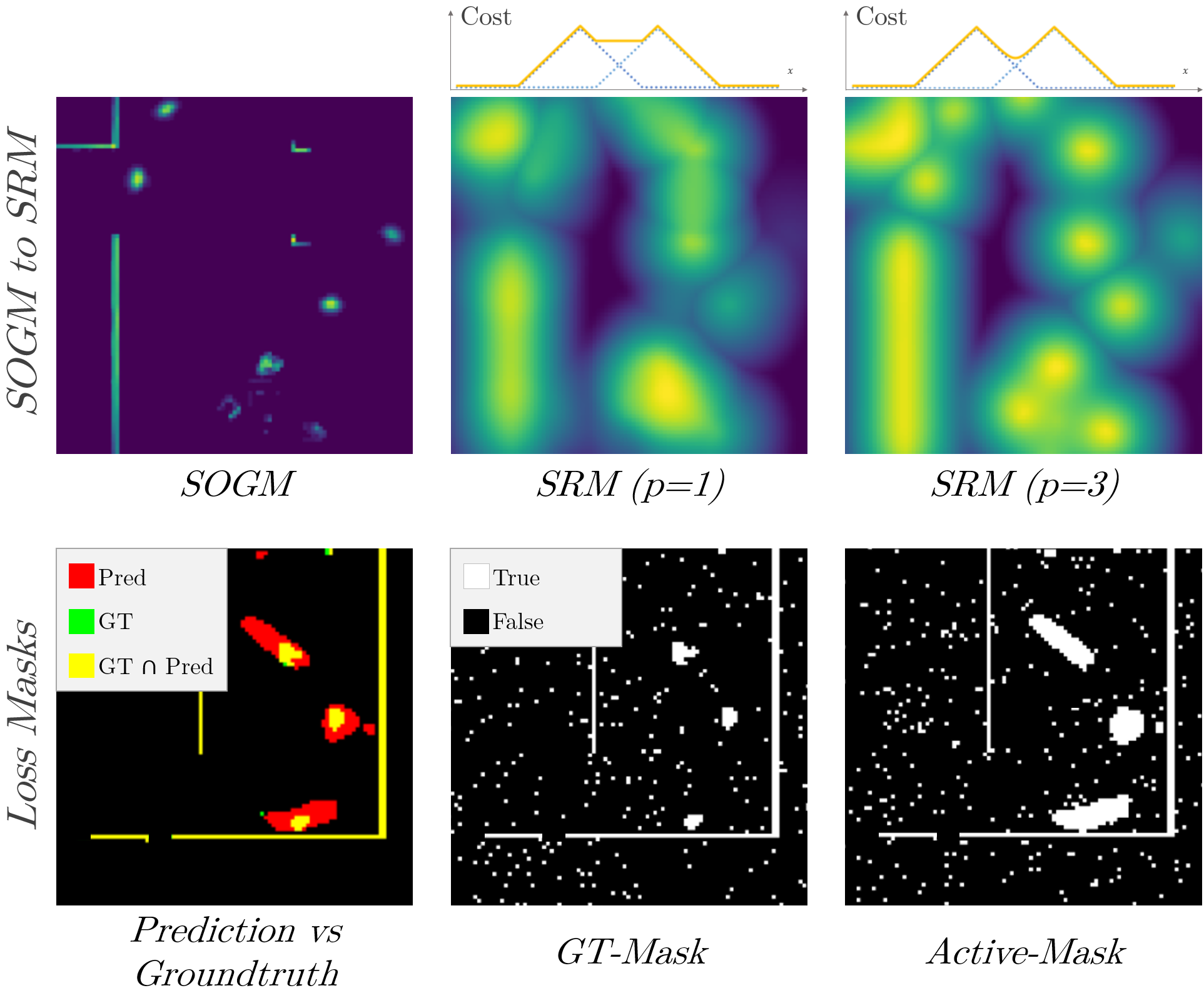}
    \caption{Top: conversion of SOGMs into SRMs for different values of $p$. Bottom: Loss masks reducing the influence of empty pixels during training.}
    \vspace{-3ex}
    \label{fig_mask}
\end{figure}

During navigation, we transform the predicted SOGMs into SRMs to encode the risk of collision instead of the occupancy probability. TEB originally minimizes a linearly decreasing cost function: $\mathcal{C}_\mathrm{obst} = \mathrm{max}\left(0, 1 - d/d_0\right)$, where $d$ is the distance to the closest obstacle and $d_0$ a predefined influence distance. We define a linearly decreasing risk value similarly:

\begin{equation}
\label{eq4}
    \mathrm{SRM}_{k, i} =  \sum\limits_{j}{\left(\left(\mathcal{C}_{i,j}\right)^p \times \mathrm{SOGM}_{k, j}\right)^{1/p}}
\end{equation}

\noindent with $\mathcal{C}_\mathrm{pix}\left(i, j\right) = \mathrm{max}\left(0, 1 -  d\left(i, j\right) \times dl_\mathrm{2D} / d_0 \right)$, where $d\left(i, j\right)$ is the distance from pixel $i$ to pixel $j$ in the grid space, and $d_0$ is set to 2m. This risk value behaves like a p-norm (see top of Figure \ref{fig_mask}), the higher $p$ is, the closer it is to the maximum value of the linear influence of each surrounding pixel. We use $p = 3$ in the following, compute a SRM for each SOGM channel and keep the pixel-wise maximum. TEB optimizes a trajectory in continuous space, so we use a trilinear interpolation of the SRM to get continuous risk values. If a pose time is too far in the future, it is ignored. TEB also allows the optimization of multiple trajectories for different homotopy classes. We keep this feature by creating estimated point obstacles at local maxima in the SOGM, ignored by the trajectory optimizer, but used for homotopy class computation.

In addition, our network predicts labels for the frame points, that we use for global planning and localization. We exclude the \textit{dynamic} points to update the global planner 2D costmap, and the \textit{movable} and \textit{dynamic} points for the localization 3D point map. Because of the network delay, we only perform update operations with these labeled frames, and use raw lidar frames instead to perform localization. On an Nvidia RTX 3090, our network forward pass takes less than $50$ms with a CPU preprocessing of around $200$ms. The SRM conversion takes about $30$ms, if we add smaller other delays (communication for example), we end up with the first 3 or 4 layers of the SRM (out of 31) being obsolete, which is acceptable.


\section{Experiments}
\label{sec:4}

\subsection{Experimental Setup}

We evaluate our method in a Gazebo simulated space\footnote{We plan on doing real-world experiments in crowded indoor spaces, but it was impossible until now due to covid-19 restrictions}, and we define three actor behaviors. \textbf{Bouncers} are more like marbles than humans, walking in straight lines and bouncing on every obstacle. \textbf{Wanderers} walk with a constant speed but randomly change direction. They also bounce off obstacles and are repelled by the robot. Finally, \textbf{Flow Followers} are driven to a goal by a precomputed force field. They randomly choose their goal in a predefined set and only try to avoid the robot when it is very close to them.

For the training of our network, we create three different datasets, one of each behavior, and use randomized session setups. For each training set, we perform $16$ sessions, including $3$ for validation, with a random population, object count, and robot tour. At test time, we can choose the robot objective and the population depending on the need of the experiment. We focus this section on our major contributions and evaluate the SOGM predictions. We provide an anecdotal example of our navigation system using SRMs in the conclusion and in the \textbf{supplementary video}\footnote{\vspace{-1ex}\urlstyle{sf}\textcolor{mypink}{\url{https://huguesthomas.github.io/icra_2022_video.html}}}.

\subsection{Quantitative Evaluation of the Network Predictions}

\begin{figure}[t]
    \centering
    \hspace*{-0.4cm} 
    \includegraphics[width=0.9\columnwidth, keepaspectratio=true]{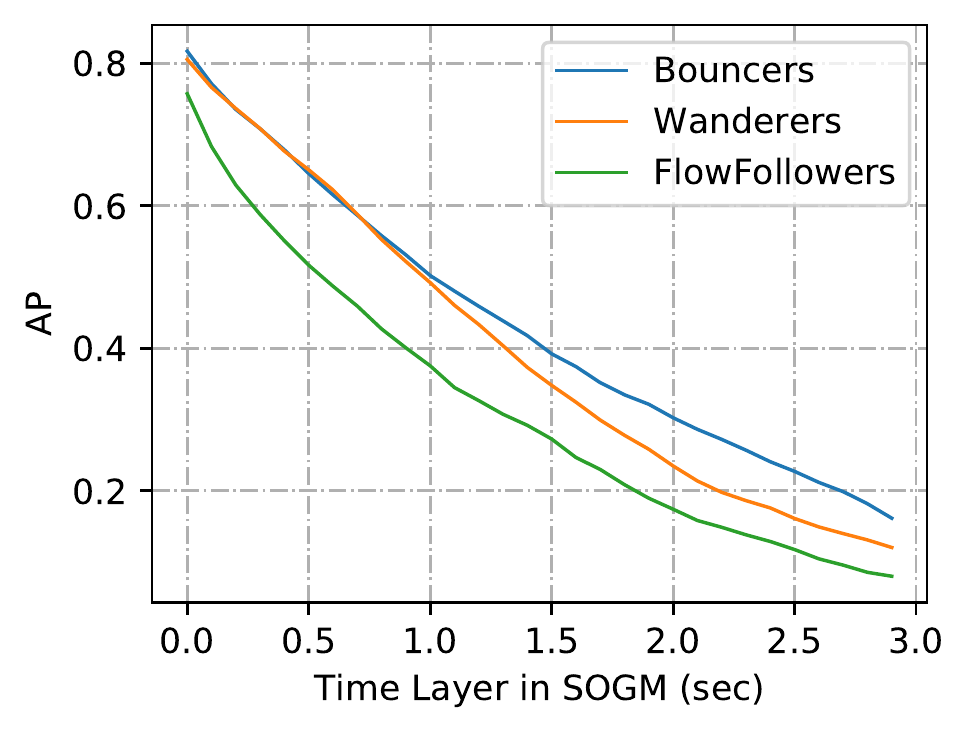}
    \vspace{-2ex}
    \caption{Comparison of the performance for our three actor behaviors.}
    \vspace{-2ex}
    \label{fig_res}
\end{figure}

We start with a quantitative analysis of the network results, and choose to only evaluate the \textit{dynamic} class performances, which are the most significant. For consistency with previous works \cite{wang2019memory, toyungyernsub2020double}, we could use Mean Squared Error (MSE) as a metric, but given the overwhelming majority of negative pixels, it is not the best choice. The SOGM predictions are 3D grids with values in $\left[0, 1\right]$, and groundtruth SOGM are binary, so we can compute precision-recall curves and use Average Precision (AP) as our main metric.

First, Figure \ref{fig_res} shows a comparison of the predicted SOGMs for Bouncers, Wanderers, and Flow Followers. By plotting AP for each time step in the SOGM, we see that it is very difficult to forecast obstacle motion far into the future. Bouncers are the easiest as they walk straight and Flow Followers are the most difficult as their movements are more complex.

We choose the Flow Followers, closer to real actors, for the following evaluations shown in Table \ref{Table_ablation}. We use the overall AP and MSE of the 3D SOGM, in addition to the AP at $1.0$s and $2.0$s. We first study different architectures for the 2D front-end network, defined by ($n_1$,$n_2$,$n_3$), respectively the number of initial levels, the number of ResNet layers per level, and the number of ResNet layers per propagation block. We note that a deeper architecture has better performances, but is also slower than our (3,2,2) architecture, chosen for fast inference. Future works could explore different front-end network designs, as there seems to be an opportunity for improvement.

\begin{table}[b]
\vspace{-3ex}
\centering
\caption{Our network outperforms the state of the art in terms of Mean Squared Error and Average Precision (at given times and overall) on the Flow Followers dataset.}
\setlength\tabcolsep{0.5pt}
\begin{footnotesize}
\begin{center}
\begin{tabular}{ | C{0.4cm} | L{2.5cm} | *{2}{C{1.2cm}} | C{1.2cm}  |  C{1.2cm} |}
\cline{3-6}
\multicolumn{2}{c|}{ } & AP$_{1.0}$ & AP$_{2.0}$ & AP$_{\mathrm{tot}}$ & MSE \TBstrut\\
\hline
\multirow{4}{*}{\rot{$\:\:\:$Ours}} & $\:\:$3D-2D (4,4,3)	& $\mathbf{31.0\%}$	& $\mathbf{11.5\%}$	& $\mathbf{27.9\%}$ & $\mathbf{3.72}$	\Tstrut\\
 & $\:\:$3D-2D (3,2,2)*	& $28.1\%$	& $8.9\%$	& $26.1\%$ &  $3.77$  \\
 & $\:\:$3D-2D (2,2,1)	& $23.7\%$	& $6.7\%$	& $23.8\%$ & $3.81$	\Bstrut\\
\hline
\hline
\multirow{4}{*}{\rot{Ablation$\:\:\,$}} & $\:\:$GT-Mask	    & $25.6\%$	& $7.5\%$	& $24.5\%$ & $3.94$	\Tstrut\\
 & $\:\:$No Mask	    & $25.7\%$	& $7.1\%$	& $23.0\%$ & $3.86$	\Bstrut\\
\cline{2-6}
 & $\:\:$no-3D-loss	  & $26.5\%$	& $8.0\%$	& $25.4\%$ & $3.83$	\Tstrut\\
 & $\:\:$shared-weights  & $3.6\%$	    & $3.1\%$	& $7.9\%$ & $20.85$	\Bstrut\\
\hline
\hline
\multirow{2}{*}{\rot{$\,$SotA}} & $\:\:$MIM-64 \cite{wang2019memory}	  & $15.9\%$	& $5.3\%$& $19.1\%$ & $41.59$	\Tstrut\\
 & $\:\:$MIM-94  & $20.7\%$	    & $7.3\%$	& $22.2\%$ & $31.04$	\Bstrut\\

\hline
\end{tabular}
\end{center}
\end{footnotesize}
\label{Table_ablation}
\vspace{-2ex}
\end{table}

We then show that the Active-Mask in our loss function is relevant, by comparing with the GT-Mask and no mask, both giving worse performances. We also show the importance of the 3D back-end to create rich features, by removing the 3D part of the loss ($\lambda_1=0$). These two experiments validate two crucial choices in our design. Eventually, we see that sharing the weights of the propagation blocks (making them recurrent) does not work. A direction for future work could be to explore recurrent convolutional layers, instead of simply sharing the weights of our propagation blocks.

Eventually, we compare with a state-of-the-art video prediction method, Memory-in-Memory (MIM) \cite{wang2019memory}, which offers a solid benchmark. We also tried the open-source implementation of \cite{toyungyernsub2020double}, with very poor results. Their method is not applicable in our case because it predicts obstacles in the robot coordinates instead of the world coordinates. MIM normally predicts $64$-pixel frames (MIM-$64$), but we modified their code to predict $94$-pixel frames (MIM-$94$) for comparable performances. We see that our 3D-2D feedforward network outperforms MIM with a large gap, proving that our performance is better than the current best comparable approaches. This validates that our major contribution, incorporating a 3D back-end to 2D OGM prediction, leads to improved performances. We share the data and parameters used in this experiment to encourage future works to explore this idea further.

\subsection{Qualitative Evaluation of the Network Predictions}

Using only numerical results, it is hard to judge how well our approach performs, and set a bar for `good' performances. Especially because the multi-modal nature of the actor trajectories lowers these values. Consider a simple example: if a person faces three doors and chooses randomly where to go, we expect the network to predict that each trajectory is possible with a $33\%$ probability, which automatically means at least $66\%$ of false positives. In our case, the situations are more complex but the principle remains, which is why we need a thorough investigation of the qualitative results to assess the performances

\begin{figure*}[t]
    \centering
    \includegraphics[width=0.999\textwidth, keepaspectratio=true]{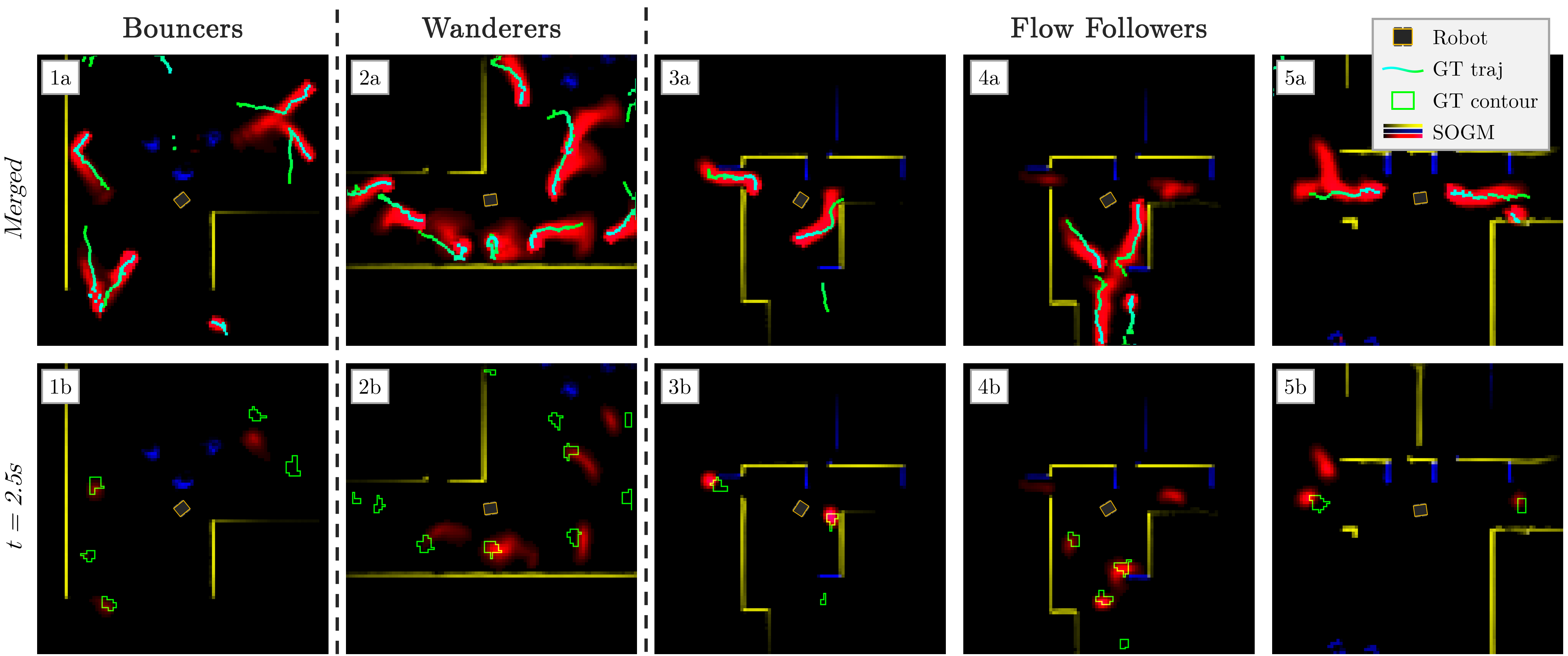}
    \vspace{-3.5ex}
    \caption{Examples of SOGM predictions. Time-step layers are merged in the top row to see temporal evolution, with superimposed groundtruth trajectories from cyan (start) to green (end). Bottom row shows a single time-step layer, with superimposed groundtruth in green.}
    \label{fig_imgs}
    \vspace{-2ex}
\end{figure*}

\begin{figure}[b]    
    \vspace{-3ex}
    \centering
    \includegraphics[width=0.999\columnwidth, keepaspectratio=true]{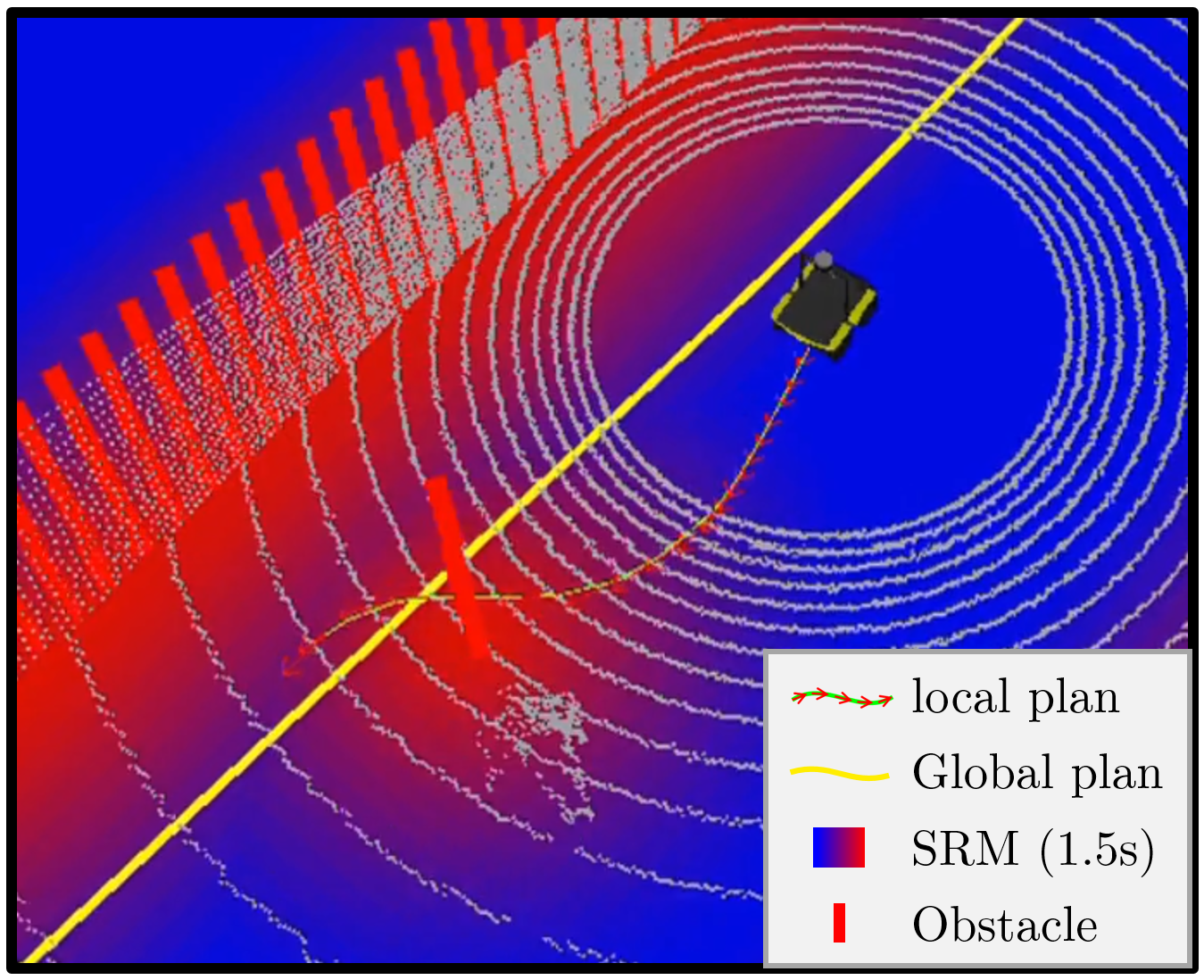}
    \caption{Proposed risk-aware navigation system. The local planner anticipates the actors' future motion and plans to go behind them.}
    \label{fig_teb}
\end{figure}

SOGMs are best seen when animated in the \textbf{supplementary video}. Here, we present them in image format (Figure \ref{fig_imgs}). First, we merged all the time steps of the SOGMs to highlight the movement that is predicted (a). We superimpose the groundtruth for the dynamic class as a trajectory, computed with a local maximum filter and coloured according to time, from cyan ($t=0.0$s) to green ($t=3.0$s). We also show one layer of prediction $2.5$ seconds into the future (b), with the groundtruth for the dynamic class superimposed as a contour. 

We can see several exciting results. First, the network handles Bouncers pretty well with accurate predictions even when they bounce on walls (1). It does not manage to predict bounces between actors, which is not surprising as they are less frequent and more random than wall bounces. Then we notice that the predictions for Wanderers (2b) generally have a shape similar to the banana distribution \cite{long2013banana}. The network is also able to predict that the Wanderers are repelled by the robot (2a). Flow Followers exhibit more complex trajectories (e.g. going through doors), that the network is able to predict (3). We observe an interesting phenomenon: the network often predicts a latent risk at some busy doorways (4), because during training the network often sees a person appearing there out of nowhere. Last but not least, our network sometimes predicts two possible trajectories, here one going through the door and one going straight (5), illustrating our ability to make multi-modal predictions.


\section{Conclusion and Future Work}
\label{sec:5}

We presented a self-supervised approach forecasting the motion of dynamic obstacles, without the need to introduce object-level description. We showed how to generate SOGMs automatically and train a novel 3D-2D feedforward neural network to predict them. Eventually, we validated our SOGM predictions both quantitatively and qualitatively. We also presented a navigation system using the predicted SOGMs to proactively avoid dynamic obstacles. We reserve a thorough evaluation of this system for future work, especially when real-world experiments will be possible again (hopefully very soon). For now, we show it running on a simple example in the supplementary video and Figure \ref{fig_teb}.

Following other works, we showed how self-supervision and lifelong learning are valuable for robotic navigation and hope that more work follows in this direction. There are many prospects for improvement, with more complex simulated human behaviors, and eventually real behaviors, by conditioning the SOGM predictions on the possible future robot actions, by combining recurrent and feed-forward architectures, or by including the notion of visible space in the generation and prediction of SOGMs.

\addtolength{\textheight}{-2cm}   


\newpage

\bibliographystyle{IEEEtran}
\bibliography{ICRA2018}

\end{document}